\pdfoutput=1

\documentclass[11pt]{article}
\usepackage[]{acl}

\usepackage[compact]{titlesec}
\usepackage{xcolor}

\usepackage{times}
\usepackage{latexsym}
\usepackage{graphicx}
\usepackage{soul}

\usepackage[T1]{fontenc}

\usepackage[utf8]{inputenc}

\usepackage{microtype}

\usepackage{inconsolata}

\title{Backprompting: Leveraging Synthetic Production Data for Health Advice Guardrails}

\author{Kellen Tan Cheng\textsuperscript{$\clubsuit \spadesuit$}\thanks{\ \  Corresponding author.}, Anna Lisa Gentile\textsuperscript{$\clubsuit$}, Chad DeLuca\textsuperscript{$\clubsuit$}, Guang-Jie Ren\textsuperscript{$\diamondsuit$}\thanks{\ \  Work done while employed at IBM Research.} \\ \textsuperscript{$\clubsuit$}IBM Research - Almaden,  
\textsuperscript{$\spadesuit$}Princeton University, 
\textsuperscript{$\diamondsuit$}Adobe Research \\
\textsuperscript{$\clubsuit$}\texttt{delucac@us.ibm.edu, annalisa.gentile@ibm.com} \\
\textsuperscript{$\spadesuit$}\texttt{kellentan@princeton.edu} 
\textsuperscript{$\diamondsuit$}\texttt{gren@adobe.com} \\
}

\definecolor{blue-violet}{rgb}{0.54, 0.17, 0.89}
\definecolor{aureolin}{rgb}{0.99, 0.93, 0.0}
\definecolor{ashgrey}{rgb}{0.7, 0.75, 0.71}
\definecolor{asparagus}{rgb}{0.53, 0.66, 0.42}
\definecolor{bubblegum}{rgb}{0.99, 0.76, 0.8}
\definecolor{byzantine}{rgb}{0.74, 0.2, 0.64}

\begin{document}
\maketitle

\begin{abstract}
The pervasiveness of large language models (LLMs) in enterprise settings has also brought forth a significant amount of risks associated with their usage. Guardrails technologies aim to mitigate this risk by filtering LLMs' input/output text through various detectors. However, developing and maintaining robust detectors faces many challenges, one of which is the difficulty in acquiring production-quality labeled data on real LLM outputs prior to deployment. In this work, we propose backprompting, a simple yet intuitive solution to generate production-like labeled data for health advice guardrails development. Furthermore, we pair our backprompting method with a sparse human-in-the-loop clustering technique to label the generated data. Our aim is to construct a parallel corpus roughly representative of the original dataset yet resembling real LLM output. We then infuse existing datasets with our synthetic examples to produce robust training data for our detector. We test our technique in one of the most difficult and nuanced guardrails: the identification of health advice in LLM output, and demonstrate improvement versus other solutions. Our detector is able to outperform GPT-4o by up to 3.73\%, despite having 400x less parameters. 
\end{abstract}

\section{Introduction}\label{sec:introduction}

The advancement of large language models (LLMs) has brought about impressive capabilities in a wide variety of natural language tasks~\cite{GPT4.all, llama3.1.all}. However, the fact that these models are pre-trained on massive text corpora inevitably results in the generation of some undesirable outputs that may be misleading and/or factually incorrect. Many prominent LLMs have methods in place to safeguard their interactions with users~\cite{nemo, llama-guard, openai-guardrails, salem2023maatphorautomatedvariantanalysis, guardrails-position}, but developing guardrails technology that can effectively minimize LLMs' usage risks remains an open challenge. Additionally, conventional techniques typically involve a nontrivial human component, whether for crafting/curating specific datasets for the task or for red-teaming. 

One prominent challenge in guardrail development is obtaining high-quality production data. This is because there exists a significant distribution shift between open-source fine-tuning (FT) datasets, which are typically human-curated, and the data that is actually encountered during inference, which is machine-generated~\cite{ibm-detectors, wilds, gradient-detection, distribution-shift-images}. Additionally, only a select few corporations have access to large-scale production datasets containing LLMs' prompts and responses, but given their proprietary nature, it is impossible to utilize them for guardrails development. This scarcity is exacerbated in domains such as healthcare or finance, due to privacy concerns and the involvement of critical decision-making within the data~\cite{distribution-shift-health, financial-gpt, finguard}. While there have been considerable efforts to construct various LLM risk benchmark datasets~\cite{red-teaming-lms, data-perf, beaver-tails, do-not-answer}, there is still a nontrivial human cost, and manually constructing benchmarks for each particular risk category is not scalable.

Towards addressing this problem, we introduce a simple yet intuitive framework for synthetic generation of real-world production data. Inspired by the concept of backtranslation~\cite{backtranslation}, backprompting uses an initial set of annotated data and generates a completely new set of data. The process begins by (i) generating a prompt for each given input text, then (ii) feeding the prompts back to an LLM, (iii) using the generated text as the new text, and (iv) performing a sparse human-in-the-loop (sparse-HITL) labeling scheme, making minimal use of human feedback to effectively produce labeled synthetic data. Our framework makes no restrictions on the type of guardrails task, the type of input data, or even the language model which is used for the text generation. 
The contributions of this work are as follows:

\noindent(1) a framework to synthetically generate data in the style of an LLM's outputs 

\noindent(2) a semi-automatic sparse-HITL annotation scheme to label the synthetically generated data

\noindent(3) a two-stage FT setup to better adapt language models towards the synthetic data, where the first stage incorporates a mix of synthetic and open-source data, and the second stage uses purely synthetic data.

\section{Related Work}\label{sec:related_work}

\subsection{Query Generation}
There have been various approaches to generate queries from input texts~\cite{qa-attention}. Prior work has focused on validating summary quality with questions~\cite{qa-summary}, generating question-answer pairs~\cite{squash}, automatic query generation for event-extraction~\cite{ee-qa}, or using templates or knowledge graphs~\cite{isq, multihop-qa, rnn-kg, template-qg}. Unlike prior work, our approach considers both query generation and the corresponding output generation, in a manner akin to backtranslation. Additionally, it is possible to complement our framework with a more optimized query generation scheme, such that it works in tandem with prior work.

\subsection{Synthetic Data Generation}
Synthetic data generation remains a pertinent and useful capability of present-day LLMs~\cite{synthetic-generation-survey, synthetic-health}. Mainstream techniques for LLMs mainly focus on a variety of prompt-engineering techniques such as creating roleplay~\cite{synthetic-roleplay}, defining task specifications or taxonomies~\cite{synthetic-mixup, instructlab}, knowledge graphs~\cite{synthetic-kg}, feedback~\cite{progen}, and in-context learning examples~\cite{self-instruct, self-prompting}. Our approach differs in that we are optimizing to match the LLM outputs' distribution, rather than data quality itself. This is because we consider data generation for the application of guardrails development, and such erroneous or dirty samples are texts that could realistically be generated by an LLM during inference. As a result, including some imperfect samples allows our detector model to be even more robust. 

\subsection{Guardrails Development}
In recent years, guardrails development for LLMs has been a prominent subfield within natural language processing~\cite{guardrails-position}. There have been a variety of different approaches to implementing guardrails, from using lightweight detector models~\cite{ibm-detectors, oneshield}, taxonomies and/or red-teaming with LLMs~\cite{llama-guard, openai-guardrails}, human programmable guardrails~\cite{nemo}, to query-modification and fusion models~\cite{rigorllm, guardagent}. Our approach is also one method for guardrails development, but instead we focus more on the data creation step, namely generating high-quality production-like data that can be used to make robust datasets for creating guardrails models. In this sense, we are less focused on an actual model framework and more on how to provide the tools (i.e. data) necessary to help facilitate guardrails development.

\section{Synthetic Data Generation}\label{sec:method}
\begin{figure*}[htbp]
\centering
\includegraphics[width=\textwidth]{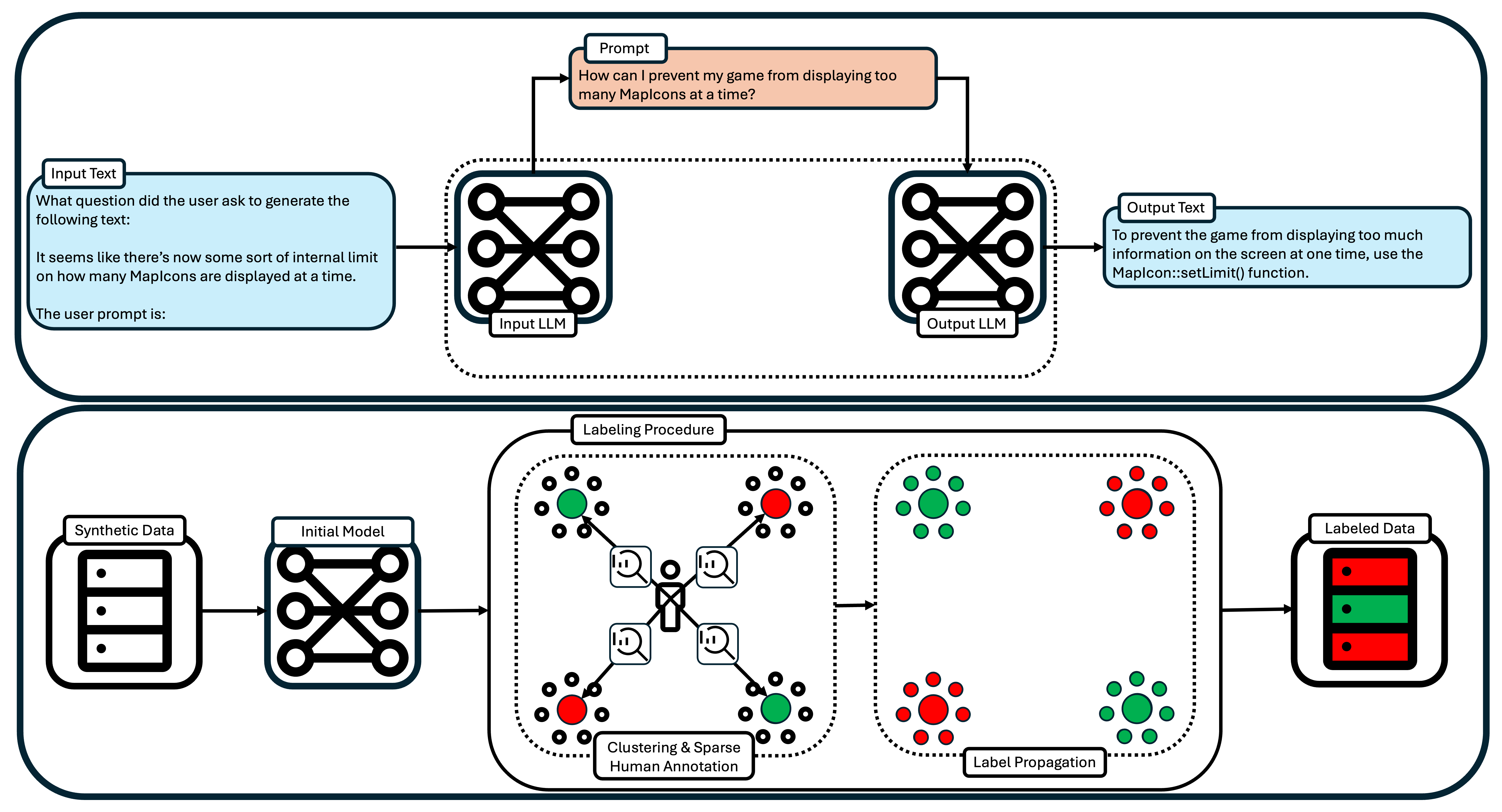}
\caption{An overview of the data generation schematic. (Top) For each data point, we first transform it into a query, and then re-prompt an LLM with our formulated query to generate a new synthetic data point. We note that for our particular implementation, the input LLM and the output LLM are the same. (Bottom) To label our synthetically generated data, we first use the initial model to split the data by predicted label, and then within each split, cluster the samples by their model embeddings. Then, a human annotator labels only the cluster centroids, before then propagating this label onto all cluster members.}
\label{fig:framework}
\end{figure*}

We describe the synthetic data generation framework seen in Figure~\ref{fig:framework}. Please refer to Appendix~\ref{sec:appendix_hyperparameters} for the comprehensive list of hyperparameters.

\subsection{Backprompting}\label{sec:method_dataGen}
Our backprompting protocol works in two stages: (i) \textit{query generation} from the original texts, and (ii) \textit{answer generation} which produces new output texts from our generated queries. 

We first take as input a seed dataset $\mathcal{X} = \{x_1, \dots, x_n\}$, where each $x_i$ is a text sample, such as a sentence, a paragraph, or a document. Since our framework is capable of automatically generating labels (Section~\ref{sec:method_hitl}), we remark that there is no restriction that the seed dataset $\mathcal{X}$ be annotated. In the \textit{query generation} stage, we produce a query set $\mathcal{Q} = \{q_1, \dots, q_n\}$ such that each query $q_i \in \mathcal{Q}$ corresponds to the text $x_i \in \mathcal{X}$, and is generated by asking our LLM which question has the text $x_i$ as a potential answer. The specific prompt template can be found in Appendix~\ref{sec:appendix_examples}.

In the \textit{answer generation} stage, we use the set of queries $\mathcal{Q}$ to generate synthetic data. For each query $q_i \in \mathcal{Q}$, we prompt our LLM to generate a response $y_i$, resulting in a synthetic dataset $\mathcal{Y} = \{y_1, \dots, y_n\}$. This data is production-quality, since each output $y_i$ is LLM-generated, and thus distributed accordingly to what would be observed in deployment settings.

\subsection{Sparse-HITL}\label{sec:method_hitl}
Even if the original dataset $\mathcal{X}$ is labeled, we cannot assume that the original label for $x_i$ holds for the synthetically generated $y_i$ -- backprompting does not guarantee complete equivalence between $x_i$ and $y_i$. Therefore, we propose sparse-HITL to perform semi-automatic annotation on our synthetic dataset $\mathcal{Y}$ without incurring high manual labor overhead. First, we use a model $\mathcal{M}$ that was fine-tuned towards the target classification task, specifically fine-tuned on the same seed datasets from which the synthetic data is generated from. We then split $\mathcal{Y}$ according to their predicted labels, as output by $\mathcal{M}$. Within each group, we generate embeddings from $\mathcal{M}$ for each sample and then cluster them using the k-means algorithm~\cite{kmeans}. Finally, we manually annotate only the cluster centroids, then propagate this human-annotated label onto all cluster members. In this manner, human annotation is only needed for one data point per cluster, thus limiting the number of manual annotations to the total number of clusters.

\subsection{Two-Stage FT}\label{sec:method_ft}
We implement FT in two stages in order to gradually align our detector model to the guardrails task at hand. The first stage is done on a combination of synthetic and open-source datasets, where the synthetic dataset is generated from a seed dataset. This synthetic data in the first stage uses only seed examples that are negative, i.e. do not violate our guardrails task. The motivation here is that during inference, a vast majority of samples will be irrelevant and not violate any guardrails. Thus, we aim to ensure that the detector model is able to accurately classify any irrelevant samples correctly as negatives, reducing the false positive rate. Without this step, the model is more prone to errors when it encounters irrelevant samples, since otherwise they would never have been seen before during FT. 

After the first stage, the model has seen a wide range of inputs and knows how to deal with irrelevant samples. In the second stage, we FT the model on purely synthetic data in order to further tune its behavior on relevant samples. In this stage, the synthetic data uses only seed examples that are positive, in order to align the detector with examples of LLM-generated text that violate the guardrails.

\section{Experimental Setup}\label{sec:task}
\subsection{Datasets}
In this work, we focus specifically on the task of health advice recognition, i.e. detecting whether a given LLM output text contains health advice. Note that we define health advice as follows: \textit{health advice (boolean) refers to any text that contains an explicit recommendation or suggestion on a course of actions that a person should take}. Importantly, this guardrails task is not concerned with distinguishing between helpful versus harmful advice, but simply whether it is present. We formulate this problem as a three-way classification task, where our labels are \textit{health-advice}, \textit{not health-advice}, and \textit{general-content}. The addition of a \textit{general-content} class helps introduce an additional layer of granularity during fine-tuning, ensuring that the predictions remain consistent for text that is not health-related. However, during inference, we treat both \textit{general-content} and \textit{not health-advice} equally as part of the negative class. Results for this task are evaluated on the gold-standard HeAL benchmark dataset~\cite{heal}\footnote{\url{https://doi.org/10.6084/m9.figshare.27198735}}.

To construct our fine-tuning dataset for stage one, we first synthetically generate general content samples using SemEval2019-Task9~\cite{semeval2019-task9} as our seed dataset. We then perform semi-automatic annotation using the sparse-HITL labeling scheme (as detailed in Section~\ref{sec:method_hitl}). We combine this synthetic data with HealthE~\cite{healthe} and Detecting-Health-Advice~\cite{detecting-health-advice}, two health advice datasets, to obtain the final stage-one fine-tuning dataset. Note that for the Detecting-Health-Advice dataset we consider both the weak advice and strong advice labeled samples as health advice.

The stage-two fine-tuning dataset is constructed from purely synthetic examples generated by the backprompting framework. We combine both HealthE and Detecting-Health-Advice, and extract all the positive samples to use as our seed data points. Note that the stage-two fine-tuning dataset contains purely synthetic data. For detailed dataset statistics, please see Appendix~\ref{sec:appendix_datasets}.

\subsection{Models}
We use Llama-3.1-8B-Instruct~\cite{llama3.1.all} as the LLM for both the query generation and answer generation stages. The sparse-HITL step utilizes a BART-Large model~\cite{bart} that was fine-tuned on Detecting-Health-Advice, HealthE, and SemEval2019-Task9 as the model $\mathcal{M}$. Clustering of the embeddings is done with $k = 20$ as our default number of clusters. Note that we use an off-the-shelf BART-Large architecture as our detector model.

\section{Results \& Discussion}\label{sec:results}
\subsection{Performance on HeAL}\label{sec:results_ft}
\begin{table*}[ht]
\begin{center}
\small
\begin{tabular}{|c|c|c|c|c|}
\hline
FT Setup & Accuracy & Precision & Recall & F1 \\
\hline\hline 
Real & 81.34\% & 86.73\% & 81.33\% & 83.94\% \\
Synthetic & 82.09\% & \textbf{92.46}\% & 76.35\% & 83.64\% \\
Real $\oplus$ Synthetic & 80.60\% & 89.76\% & 76.35\% & 82.51\% \\
\hline
GPT-4o & 81.59\% & 79.51\% & \textbf{93.36}\% & 85.88\% \\
Llama-3-70B-Instruct & 81.34\% & 85.78\% & 82.57\% & 84.14\% \\
Mixtral-8x7B & 72.89\% & 79.15\% & 72.61\% & 75.74\% \\
\hline
Stage-1 & 75.62\% & 84.88\% & 72.20\% & 78.03\% \\
Stage-2 & \textbf{85.32}\% & 89.91\% & 85.06\% & \textbf{87.42}\% \\
Alternate Stage-2 & 81.34\% & 85.78\% & 82.57\% & 84.14\% \\
\hline
\end{tabular}
\end{center}
\caption{Performance of our detector model across different baseline and fine-tuning setups. Note that Alternate Stage-2 refers to when we instead use examples generated from positive seeds in the first stage of fine-tuning, and those generated from negative seeds in the second stage. The best-performing results are in \textbf{bold}.}
\label{tab:ft_results}
\end{table*}

\noindent\textbf{2-Stage FT} We observe in Table~\ref{tab:ft_results} that 2-stage FT with our detector model achieves state-of-the-art performance, beating out GPT-4o by 3.73\% in accuracy and 1.54\% in F1-score. This gain is statistically significant up to 90\% confidence, and is achieved despite our detector model (BART-Large) containing only 400M parameters, in stark contrast to Mixtral-8x7B~\cite{mixtral}, Llama-3-70B-Instruct~\cite{llama3.1.all}, and GPT-4o~\cite{GPT4.all}. Additionally, our detector model also exhibits relatively balanced behavior when encountering negative versus positive samples, evidenced by a difference of just 4.85\% between its precision and recall. While this is slightly larger than Llama-3-70B-Instruct (3.21\%), it is still a vast improvement compared to GPT-4o with 13.85\%. Large differences between precision and recall are undesirable in deployment settings, since they result in detectors that are too skewed one way or another.

Furthermore, we note that our results also demonstrate why we only use synthetic samples corresponding to positive seeds in the second stage of fine-tuning, as opposed to the first stage. From Table~\ref{tab:ft_results}, using positive seeds in the first stage and negative seeds in the second stage (Alternate Stage-2) degrade the detector performance, achieving only 81.34\% accuracy and 84.14\% in F1-score. While these results are still comparable with Llama-3-70B-Instruct, it exhibits a drop of 3.98\% accuracy and 3.28\% in F1-score compared to 2-stage FT. 

\noindent\textbf{1-Stage FT} Interestingly, even the addition of just synthetic irrelevant samples (general content) can result in a performance improvement. From Table~\ref{tab:ft_results}, fine-tuning with just the first stage is already sufficient to outperform a much larger model such as Mixtral-8x7B, with an improvement of 2.73\% in accuracy and 2.29\% in F1-score. However, 1-stage FT is unable to outperform GPT-4o or Llama-3-70B-Instruct on its own, and its performance also lags behind vanilla FT approaches in all settings. The performance gains with 2-stage FT, which outperforms 1-stage FT by 9.70\% in accuracy and 9.39\% in F1-score, demonstrate its importance to better improve the model performance and balance out the detector behavior.

\noindent\textbf{Vanilla FT} We also compare the performance of vanilla fine-tuning using only real datasets, synthetic datasets, as well as both real and synthetic datasets. From Table~\ref{tab:ft_results}, we see that fine-tuning on synthetic data versus real data yields comparable results, as the detector fine-tuned on synthetic data achieves higher accuracy (82.09\% v. 81.34\%) but slightly lower F1-score (83.64\% v. 83.94\%). However, we note that fine-tuning on synthetic data does result in a larger difference between precision and recall at 16.11\%, as opposed to 5.40\% when fine-tuned on real data. Additionally, it appears that fine-tuning on both real and synthetic data actually results in worse performance than fine-tuning on either real or synthetic data.

\subsection{Synthetic Data Quality}
Another point of interest focuses on the quality of the synthetic data. To gauge the noisiness of the sparse-HITL labels, we randomly sampled two examples from each cluster to manually annotate for label accuracy. From Table~\ref{tab:manual_annotation_scores}, all datasets exhibit at least 87.50\% accuracy, with only HealthE containing a false negative sample. Most of the erroneously labeled samples arise as false positives, where the text is indeed health or medical-related but does not contain explicit advice. 

\begin{table}[ht]
\begin{center}
\small
\begin{tabular}{|c|c|c|c|}
\hline
Dataset & P & R & F1 \\
\hline\hline 
DHA & \textbf{80.33}\% & \textbf{83.58}\% & \textbf{81.91}\% \\
HealthE & 80.23\% & 81.04\% & 80.61\% \\
SemEval & 79.44\% & 82.60\% & 80.96\% \\
\hline
\end{tabular}
\end{center}
\caption{Average BERTScore similarities between the synthetic data and their corresponding seed examples. Note that P and R denote precision and recall, respectively. DHA denotes the Detecting-Health-Advice dataset. The best performing scores are in \textbf{bold}.}
\label{tab:bertscore_similarities}
\end{table}

\begin{table}[ht]
\begin{center}
\small
\begin{tabular}{|c|c|c|c|}
\hline
Dataset & FP & FN & Accuracy\\
\hline\hline 
DHA & 5 & 0 & 87.50\% \\
HealthE & 3 & 1 & 90.00\% \\
SemEval & 0 & 0 & 100.00\% \\
\hline
\end{tabular}
\end{center}
\caption{Manual annotation of synthetic data label accuracy. Note that FP and FN stand for false positives and false negatives, respectively. DHA denotes the Detecting-Health-Advice dataset.}
\label{tab:manual_annotation_scores}
\end{table}

\begin{table*}[ht]
\begin{center}
\small
\begin{tabular}{|c|c|c|c|c|c|c|}
\hline
Dataset & HC & HA & GC & Health \% & HA \% & Cluster Size Std\\
\hline\hline 
DHA & 1461 & 496 & 790 & 71.24\% & 18.06\% & 62.41 \\
HealthE & 1455 & 1847 & 98 & 97.12\% & 54.32\% & 53.57 \\
SemEval & 0 & 0 & 9925 & 0.00\% & 0.00\% & 199.06 \\
\hline
\end{tabular}
\end{center}
\caption{Analysis of the synthetic data label distributions and cluster statistics. Note that HC, HA, and GC denote health content, health advice, and general content, respectively. DHA denotes the Detecting-Health-Advice dataset. Health \% indicates the percentage of synthetic data that is labeled as either HC or HA, while HA \% indicates the percentage of synthetic data that is labeled as HA. Finally, we also include the standard deviation of the cluster sizes within each split.}
\label{tab:synthetic_data_statistics}
\end{table*}

As a quantitative metric, we also evaluate the semantic drift between the synthetic and seed examples using BERTScore~\cite{bertscore}, a metric designed to evaluate the quality of the generated text. As evidenced from Table~\ref{tab:bertscore_similarities}, synthetic data generated from Detecting-Health-Advice exhibits the lowest semantic drift, achieving the highest BERTScore results. On the other hand, semantic drift is comparable between samples generated via HealthE versus SemEval, with HealthE achieving a slightly higher BERTScore precision of 80.23\% but a slightly lower BERTScore F1-score of 80.61\%. The slightly lower semantic drift from health advice datasets is expected: general content samples are less focused on a particular topic and thus are more likely to exhibit semantic drift from the original seed examples. This can be further seen in Table~\ref{tab:synthetic_data_statistics}, where we observe that the data generated from health advice datasets tend to stay within the health domain. Specifically, 71.24\% and 97.12\% of the synthetically generated samples are labeled as health (either health content or health advice) for Detecting-Health-Advice and HealthE, respectively. It appears that the examples generated from HealthE are more likely to also stay as health advice, with 54.32\% of the synthetic examples being labeled as health advice (in keeping with the original label), as opposed to 18.06\% for Detecting-Health-Advice. While it seems that semantic drift can push seed examples from health advice to general content, the same cannot be said in reverse, with all examples generated from SemEval still being labeled as general content.

However, as we discussed in our prior results, some degree of semantic drift is desirable. We hypothesize that this is because we expose the detector model to a wider range of LLM outputs. This wider distribution makes the detector better equipped to handle irrelevant data points, which compose a prominent part of the data it sees during inference. Additionally, some amount of dirty data also helps make the detector model robust, since real production data may also be imperfect, given that it is generated by LLMs. In these scenarios, some prior exposure helps ensure the model is not wildly inconsistent on these samples.

\subsection{Deployment}
The backprompting framework has the potential to generate production-like synthetic data for a wide variety of guardrails tasks outside of just health advice identification, and the framework is especially geared towards advice guardrails (e.g. legal advice, financial advice), where the detector needs to distinguish between domain-related text and domain advice-related text. 
This Heath Advice Detector model trained with this method has been deployed within an internal suite of guardrails for LLMs\footnote{Reference omitted for anonymity.}. The suite also includes detectors that cover risks such as personal identifiable information (PII), hate, abuse, and profanity (HAP), inappropriate content (e.g., pornography), as well as a text attribution detector designed to identify potential copyright violations and/or the leakage of proprietary information. This suite has been integrated into a product\footnote{Reference omitted for anonymity.} and is utilized in several real-world use cases and end-to-end demonstrations, including an externally-facing open-source project\footnote{Reference omitted for anonymity.}.

\section{Conclusion \& Future Work}\label{sec:conclusion}
In this work, we present an intuitive and effective framework for automating the generation of production-quality synthetic data. Backprompting functions by transforming input texts into their corresponding queries, and then feeding those queries into the LLM for text generation. We also design a sparse human-in-the-loop (sparse-HITL) clustering method to cluster the synthetically generated data, and manually annotate only the centroids (i.e. representative samples). This scheme ensures minimal use of human labor but maximizes the benefits, propagating the manually annotated label onto all data points within that cluster. We demonstrate the efficacy of our approach on one of the most difficult guardrails tasks, which is the identification of health advice in LLM outputs. Our results demonstrate that we can beat even the largest contemporary LLMs, such as GPT-4o, by up to 3.73\%, and outperform standard fine-tuning and alternative approaches on both benchmark datasets and real-world production data (see Appendix~\ref{sec:appendix_bam_evaluation}).

There are many avenues for future work, since backprompting is a highly modular framework that allows for the development of each component in isolation, before ultimately combining the methods. Improving the query generation procedure to reduce semantic drift and mitigate the amount of noisy samples is one avenue of research. Additionally, further work can improve upon our generation setup, whether it's through the use of newer models (as they arrive) or specialized text generation schema. Finally, it would be interesting to understand how our method performs in low-resource settings, where open-source datasets do not readily exist for that guardrails task. In this case, one would have to crawl web data for that guardrails domain to create the seed dataset, and then input those texts into the backprompting framework.

\section{Ethics Statement}\label{sec:ethics_statement}
In Appendix~\ref{sec:appendix_bam_evaluation}, we did report some metrics on actual, real-world internal production data. However, those models cannot be released due to the sensitive information that may be present within our internal production data. Additionally, to safeguard against exposing internal information, we use public, open-source, and peer-reviewed datasets for the data generation and evaluation of our framework, to ensure that the input data is as clean as possible, and does not contain personal medical records and other information. This means that all datasets used for backprompting are openly accessible and peer-reviewed -- evaluation on internal data was only done to demonstrate the viability of our approach when tested on actual data. 

Just like there are inherent risks present in LLMs, we recognize that no model is always safe, and each model contains their own inherent risks. As a result, we urge future users of our method to validate that the results make sense and are positive for their particular use case before deployment. For use cases which don't require the data to be distributed from an LLM's underlying distribution, then backprompting may not be fully utilized in that sense.

Finally, all of our detector models are lightweight architectures (roughly 400M parameters), relatively speaking. Additionally, we utilized all LLMs in a quantized 4-bit manner to reduce our total carbon emissions impact and improve GPU efficiency. Note that our entire framework can be executed on a single GPU, as we want backprompting to be widely available and not restricted due to excessive compute requirements. Please refer to Appendix~\ref{sec:appendix_software_model_compute} for the full details.

\bibliography{custom.bib}

\begin{thebibliography}{47}
\providecommand{\natexlab}[1]{#1}

\bibitem[{Achintalwar et~al.(2024)Achintalwar, Garcia, Anaby-Tavor, Baldini, Berger, Bhattacharjee, Bouneffouf, Chaudhury, Chen, Chiazor, Daly, DB, de~Paula, Dognin, Farchi, Ghosh, Hind, Horesh, Kour, Lee, Madaan, Mehta, Miehling, Murugesan, Nagireddy, Padhi, Piorkowski, Rawat, Raz, Sattigeri, Strobelt, Swaminathan, Tillmann, Trivedi, Varshney, Wei, Witherspooon, and Zalmanovici}]{ibm-detectors}
Swapnaja Achintalwar, Adriana~Alvarado Garcia, Ateret Anaby-Tavor, Ioana Baldini, Sara~E. Berger, Bishwaranjan Bhattacharjee, Djallel Bouneffouf, Subhajit Chaudhury, Pin-Yu Chen, Lamogha Chiazor, Elizabeth~M. Daly, Kirushikesh DB, Rogério~Abreu de~Paula, Pierre Dognin, Eitan Farchi, Soumya Ghosh, Michael Hind, Raya Horesh, George Kour, and 19 others. 2024.
\newblock \href {https://arxiv.org/abs/2403.06009} {Detectors for safe and reliable llms: Implementations, uses, and limitations}.
\newblock \emph{Preprint}, arXiv:2403.06009.

\bibitem[{Cheng et~al.(2024)Cheng, Gentile, Li, DeLuca, and Ren}]{heal}
Kellen~Tan Cheng, Anna~Lisa Gentile, Pengyuan Li, Chad DeLuca, and Guang-Jie Ren. 2024.
\newblock Don't be my doctor! recognizing healthcare advice in large language models.
\newblock In \emph{Proceedings of the 2024 Conference on Empirical Methods in Natural Language Processing: Industry Track}, Miami. Association for Computational Linguistics.

\bibitem[{DeLuca et~al.(2025)DeLuca, Gentile, Asthana, Zhang, Chowdhary, Cheng, Shbita, Li, Ren, and Gopisetty}]{oneshield}
Chad DeLuca, Anna~Lisa Gentile, Shubhi Asthana, Bing Zhang, Pawan Chowdhary, Kellen Cheng, Basel Shbita, Pengyuan Li, Guang-Jie Ren, and Sandeep Gopisetty. 2025.
\newblock \href {https://arxiv.org/abs/2507.21170} {Oneshield -- the next generation of llm guardrails}.
\newblock \emph{Preprint}, arXiv:2507.21170.

\bibitem[{Dong et~al.(2024)Dong, Mu, Jin, Qi, Hu, Zhao, Meng, Ruan, and Huang}]{guardrails-position}
Yi~Dong, Ronghui Mu, Gaojie Jin, Yi~Qi, Jinwei Hu, Xingyu Zhao, Jie Meng, Wenjie Ruan, and Xiaowei Huang. 2024.
\newblock \href {https://proceedings.mlr.press/v235/dong24c.html} {Position: Building guardrails for large language models requires systematic design}.
\newblock In \emph{Proceedings of the 41st International Conference on Machine Learning}, volume 235 of \emph{Proceedings of Machine Learning Research}, pages 11375--11394. PMLR.

\bibitem[{Du et~al.(2024)Du, Li, Zhou, Ding, Wang, Zhou, and Liu}]{finguard}
Wenlong Du, Qingquan Li, Jian Zhou, Xu~Ding, Xuewei Wang, Zhongjun Zhou, and Jin Liu. 2024.
\newblock \href {https://doi.org/10.1145/3595916.3626351} {Finguard: A multimodal aigc guardrail in financial scenarios}.
\newblock In \emph{Proceedings of the 5th ACM International Conference on Multimedia in Asia}, MMAsia '23, New York, NY, USA. Association for Computing Machinery.

\bibitem[{Du et~al.(2017)Du, Shao, and Cardie}]{qa-attention}
Xinya Du, Junru Shao, and Claire Cardie. 2017.
\newblock \href {https://doi.org/10.18653/v1/P17-1123} {Learning to ask: Neural question generation for reading comprehension}.
\newblock In \emph{Proceedings of the 55th Annual Meeting of the Association for Computational Linguistics (Volume 1: Long Papers)}, pages 1342--1352, Vancouver, Canada. Association for Computational Linguistics.

\bibitem[{Dubey et~al.(2024)Dubey, Jauhri, Pandey, Kadian, Al-Dahle, Letman, Mathur, Schelten, Yang, Fan, Goyal, Hartshorn, Yang, Mitra, Sravankumar, Korenev, Hinsvark, Rao, Zhang, Rodriguez, Gregerson, Spataru, Roziere, Biron, Tang, Chern, Caucheteux, Nayak, Bi, Marra, McConnell, Keller, Touret, Wu, Wong, Ferrer, Nikolaidis, Allonsius, Song, Pintz, Livshits, Esiobu, Choudhary, Mahajan, Garcia-Olano, Perino, Hupkes, Lakomkin, AlBadawy, Lobanova, Dinan, Smith, Radenovic, Zhang, Synnaeve, Lee, Anderson, Nail, Mialon, Pang, Cucurell, Nguyen, Korevaar, Xu, Touvron, Zarov, Ibarra, Kloumann, Misra, Evtimov, Copet, Lee, Geffert, Vranes, Park, Mahadeokar, Shah, van~der Linde, Billock, Hong, Lee, Fu, Chi, Huang, Liu, Wang, Yu, Bitton, Spisak, Park, Rocca, Johnstun, Saxe, Jia, Alwala, Upasani, Plawiak, Li, Heafield, Stone, El-Arini, Iyer, Malik, Chiu, Bhalla, Rantala-Yeary, van~der Maaten, Chen, Tan, Jenkins, Martin, Madaan, Malo, Blecher, Landzaat, de~Oliveira, Muzzi, Pasupuleti, Singh, Paluri, Kardas, Oldham, Rita,
  Pavlova, Kambadur, Lewis, Si, Singh, Hassan, Goyal, Torabi, Bashlykov, Bogoychev, Chatterji, Duchenne, Çelebi, Alrassy, Zhang, Li, Vasic, Weng, Bhargava, Dubal, Krishnan, Koura, Xu, He, Dong, Srinivasan, Ganapathy, Calderer, Cabral, Stojnic, Raileanu, Girdhar, Patel, Sauvestre, Polidoro, Sumbaly, Taylor, Silva, Hou, Wang, Hosseini, Chennabasappa, Singh, Bell, Kim, Edunov, Nie, Narang, Raparthy, Shen, Wan, Bhosale, Zhang, Vandenhende, Batra, Whitman, Sootla, Collot, Gururangan, Borodinsky, Herman, Fowler, Sheasha, Georgiou, Scialom, Speckbacher, Mihaylov, Xiao, Karn, Goswami, Gupta, Ramanathan, Kerkez, Gonguet, Do, Vogeti, Petrovic, Chu, Xiong, Fu, Meers, Martinet, Wang, Tan, Xie, Jia, Wang, Goldschlag, Gaur, Babaei, Wen, Song, Zhang, Li, Mao, Coudert, Yan, Chen, Papakipos, Singh, Grattafiori, Jain, Kelsey, Shajnfeld, Gangidi, Victoria, Goldstand, Menon, Sharma, Boesenberg, Vaughan, Baevski, Feinstein, Kallet, Sangani, Yunus, Lupu, Alvarado, Caples, Gu, Ho, Poulton, Ryan, Ramchandani, Franco, Saraf,
  Chowdhury, Gabriel, Bharambe, Eisenman, Yazdan, James, Maurer, Leonhardi, Huang, Loyd, Paola, Paranjape, Liu, Wu, Ni, Hancock, Wasti, Spence, Stojkovic, Gamido, Montalvo, Parker, Burton, Mejia, Wang, Kim, Zhou, Hu, Chu, Cai, Tindal, Feichtenhofer, Civin, Beaty, Kreymer, Li, Wyatt, Adkins, Xu, Testuggine, David, Parikh, Liskovich, Foss, Wang, Le, Holland, Dowling, Jamil, Montgomery, Presani, Hahn, Wood, Brinkman, Arcaute, Dunbar, Smothers, Sun, Kreuk, Tian, Ozgenel, Caggioni, Guzmán, Kanayet, Seide, Florez, Schwarz, Badeer, Swee, Halpern, Thattai, Herman, Sizov, Guangyi, Zhang, Lakshminarayanan, Shojanazeri, Zou, Wang, Zha, Habeeb, Rudolph, Suk, Aspegren, Goldman, Damlaj, Molybog, Tufanov, Veliche, Gat, Weissman, Geboski, Kohli, Asher, Gaya, Marcus, Tang, Chan, Zhen, Reizenstein, Teboul, Zhong, Jin, Yang, Cummings, Carvill, Shepard, McPhie, Torres, Ginsburg, Wang, Wu, U, Saxena, Prasad, Khandelwal, Zand, Matosich, Veeraraghavan, Michelena, Li, Huang, Chawla, Lakhotia, Huang, Chen, Garg, A, Silva, Bell,
  Zhang, Guo, Yu, Moshkovich, Wehrstedt, Khabsa, Avalani, Bhatt, Tsimpoukelli, Mankus, Hasson, Lennie, Reso, Groshev, Naumov, Lathi, Keneally, Seltzer, Valko, Restrepo, Patel, Vyatskov, Samvelyan, Clark, Macey, Wang, Hermoso, Metanat, Rastegari, Bansal, Santhanam, Parks, White, Bawa, Singhal, Egebo, Usunier, Laptev, Dong, Zhang, Cheng, Chernoguz, Hart, Salpekar, Kalinli, Kent, Parekh, Saab, Balaji, Rittner, Bontrager, Roux, Dollar, Zvyagina, Ratanchandani, Yuvraj, Liang, Alao, Rodriguez, Ayub, Murthy, Nayani, Mitra, Li, Hogan, Battey, Wang, Maheswari, Howes, Rinott, Bondu, Datta, Chugh, Hunt, Dhillon, Sidorov, Pan, Verma, Yamamoto, Ramaswamy, Lindsay, Lindsay, Feng, Lin, Zha, Shankar, Zhang, Zhang, Wang, Agarwal, Sajuyigbe, Chintala, Max, Chen, Kehoe, Satterfield, Govindaprasad, Gupta, Cho, Virk, Subramanian, Choudhury, Goldman, Remez, Glaser, Best, Kohler, Robinson, Li, Zhang, Matthews, Chou, Shaked, Vontimitta, Ajayi, Montanez, Mohan, Kumar, Mangla, Albiero, Ionescu, Poenaru, Mihailescu, Ivanov, Li, Wang,
  Jiang, Bouaziz, Constable, Tang, Wang, Wu, Wang, Xia, Wu, Gao, Chen, Hu, Jia, Qi, Li, Zhang, Zhang, Adi, Nam, Yu, Wang, Hao, Qian, He, Rait, DeVito, Rosnbrick, Wen, Yang, and Zhao}]{llama3.1.all}
Abhimanyu Dubey, Abhinav Jauhri, Abhinav Pandey, Abhishek Kadian, Ahmad Al-Dahle, Aiesha Letman, Akhil Mathur, Alan Schelten, Amy Yang, Angela Fan, Anirudh Goyal, Anthony Hartshorn, Aobo Yang, Archi Mitra, Archie Sravankumar, Artem Korenev, Arthur Hinsvark, Arun Rao, Aston Zhang, and 516 others. 2024.
\newblock \href {https://arxiv.org/abs/2407.21783} {The llama 3 herd of models}.
\newblock \emph{Preprint}, arXiv:2407.21783.

\bibitem[{Fabbri et~al.(2020)Fabbri, Ng, Wang, Nallapati, and Xiang}]{template-qg}
Alexander Fabbri, Patrick Ng, Zhiguo Wang, Ramesh Nallapati, and Bing Xiang. 2020.
\newblock \href {https://doi.org/10.18653/v1/2020.acl-main.413} {Template-based question generation from retrieved sentences for improved unsupervised question answering}.
\newblock In \emph{Proceedings of the 58th Annual Meeting of the Association for Computational Linguistics}, pages 4508--4513, Online. Association for Computational Linguistics.

\bibitem[{Ganguli et~al.(2022)Ganguli, Lovitt, Kernion, Askell, Bai, Kadavath, Mann, Perez, Schiefer, Ndousse, Jones, Bowman, Chen, Conerly, DasSarma, Drain, Elhage, El-Showk, Fort, Hatfield-Dodds, Henighan, Hernandez, Hume, Jacobson, Johnston, Kravec, Olsson, Ringer, Tran-Johnson, Amodei, Brown, Joseph, McCandlish, Olah, Kaplan, and Clark}]{red-teaming-lms}
Deep Ganguli, Liane Lovitt, Jackson Kernion, Amanda Askell, Yuntao Bai, Saurav Kadavath, Ben Mann, Ethan Perez, Nicholas Schiefer, Kamal Ndousse, Andy Jones, Sam Bowman, Anna Chen, Tom Conerly, Nova DasSarma, Dawn Drain, Nelson Elhage, Sheer El-Showk, Stanislav Fort, and 17 others. 2022.
\newblock \href {https://arxiv.org/abs/2209.07858} {Red teaming language models to reduce harms: Methods, scaling behaviors, and lessons learned}.
\newblock \emph{Preprint}, arXiv:2209.07858.

\bibitem[{Gatto et~al.(2023)Gatto, Seegmiller, M~Johnston, Basak, and Masud~Preum}]{healthe}
Joseph Gatto, Parker Seegmiller, Garrett M~Johnston, Madhusudan Basak, and Sarah Masud~Preum. 2023.
\newblock \href {https://doi.org/10.1609/icwsm.v17i1.22210} {Healthe: Recognizing health advice \& entities in online health communities}.
\newblock \emph{Proceedings of the International AAAI Conference on Web and Social Media}, 17(1):1024--1033.

\bibitem[{Gaur et~al.(2022)Gaur, Gunaratna, Srinivasan, and Jin}]{isq}
Manas Gaur, Kalpa Gunaratna, Vijay Srinivasan, and Hongxia Jin. 2022.
\newblock \href {https://doi.org/10.1609/aaai.v36i10.21312} {Iseeq: Information seeking question generation using dynamic meta-information retrieval and knowledge graphs}.
\newblock \emph{Proceedings of the AAAI Conference on Artificial Intelligence}, 36(10):10672--10680.

\bibitem[{Huang et~al.(2021)Huang, Geng, and Li}]{gradient-detection}
Rui Huang, Andrew Geng, and Yixuan Li. 2021.
\newblock \href {https://proceedings.neurips.cc/paper_files/paper/2021/file/063e26c670d07bb7c4d30e6fc69fe056-Paper.pdf} {On the importance of gradients for detecting distributional shifts in the wild}.
\newblock In \emph{Advances in Neural Information Processing Systems}, volume~34, pages 677--689. Curran Associates, Inc.

\bibitem[{Inan et~al.(2023)Inan, Upasani, Chi, Rungta, Iyer, Mao, Tontchev, Hu, Fuller, Testuggine, and Khabsa}]{llama-guard}
Hakan Inan, Kartikeya Upasani, Jianfeng Chi, Rashi Rungta, Krithika Iyer, Yuning Mao, Michael Tontchev, Qing Hu, Brian Fuller, Davide Testuggine, and Madian Khabsa. 2023.
\newblock \href {https://arxiv.org/abs/2312.06674} {Llama guard: Llm-based input-output safeguard for human-ai conversations}.
\newblock \emph{Preprint}, arXiv:2312.06674.

\bibitem[{Ji et~al.(2023)Ji, Liu, Dai, Pan, Zhang, Bian, Chen, Sun, Wang, and Yang}]{beaver-tails}
Jiaming Ji, Mickel Liu, Josef Dai, Xuehai Pan, Chi Zhang, Ce~Bian, Boyuan Chen, Ruiyang Sun, Yizhou Wang, and Yaodong Yang. 2023.
\newblock \href {https://proceedings.neurips.cc/paper_files/paper/2023/file/4dbb61cb68671edc4ca3712d70083b9f-Paper-Datasets_and_Benchmarks.pdf} {Beavertails: Towards improved safety alignment of llm via a human-preference dataset}.
\newblock In \emph{Advances in Neural Information Processing Systems}, volume~36, pages 24678--24704. Curran Associates, Inc.

\bibitem[{Jiang et~al.(2024)Jiang, Sablayrolles, Roux, Mensch, Savary, Bamford, Chaplot, de~las Casas, Hanna, Bressand, Lengyel, Bour, Lample, Lavaud, Saulnier, Lachaux, Stock, Subramanian, Yang, Antoniak, Scao, Gervet, Lavril, Wang, Lacroix, and Sayed}]{mixtral}
Albert~Q. Jiang, Alexandre Sablayrolles, Antoine Roux, Arthur Mensch, Blanche Savary, Chris Bamford, Devendra~Singh Chaplot, Diego de~las Casas, Emma~Bou Hanna, Florian Bressand, Gianna Lengyel, Guillaume Bour, Guillaume Lample, Lélio~Renard Lavaud, Lucile Saulnier, Marie-Anne Lachaux, Pierre Stock, Sandeep Subramanian, Sophia Yang, and 7 others. 2024.
\newblock \href {https://arxiv.org/abs/2401.04088} {Mixtral of experts}.
\newblock \emph{Preprint}, arXiv:2401.04088.

\bibitem[{Koh et~al.(2021)Koh, Sagawa, Marklund, Xie, Zhang, Balsubramani, Hu, Yasunaga, Phillips, Gao, Lee, David, Stavness, Guo, Earnshaw, Haque, Beery, Leskovec, Kundaje, Pierson, Levine, Finn, and Liang}]{wilds}
Pang~Wei Koh, Shiori Sagawa, Henrik Marklund, Sang~Michael Xie, Marvin Zhang, Akshay Balsubramani, Weihua Hu, Michihiro Yasunaga, Richard~Lanas Phillips, Irena Gao, Tony Lee, Etienne David, Ian Stavness, Wei Guo, Berton Earnshaw, Imran Haque, Sara~M Beery, Jure Leskovec, Anshul Kundaje, and 4 others. 2021.
\newblock \href {https://proceedings.mlr.press/v139/koh21a.html} {Wilds: A benchmark of in-the-wild distribution shifts}.
\newblock In \emph{Proceedings of the 38th International Conference on Machine Learning}, volume 139 of \emph{Proceedings of Machine Learning Research}, pages 5637--5664. PMLR.

\bibitem[{Krishna and Iyyer(2019)}]{squash}
Kalpesh Krishna and Mohit Iyyer. 2019.
\newblock \href {https://doi.org/10.18653/v1/P19-1224} {Generating question-answer hierarchies}.
\newblock In \emph{Proceedings of the 57th Annual Meeting of the Association for Computational Linguistics}, pages 2321--2334, Florence, Italy. Association for Computational Linguistics.

\bibitem[{Kruschwitz and Schmidhuber(2024)}]{synthetic-health}
Udo Kruschwitz and Maximilian Schmidhuber. 2024.
\newblock \href {https://aclanthology.org/2024.trac-1.6} {{LLM}-based synthetic datasets: Applications and limitations in toxicity detection}.
\newblock In \emph{Proceedings of the Fourth Workshop on Threat, Aggression {\&} Cyberbullying @ LREC-COLING-2024}, pages 37--51, Torino, Italia. ELRA and ICCL.

\bibitem[{Kumar et~al.(2019)Kumar, Hua, Ramakrishnan, Qi, Gao, and Li}]{multihop-qa}
Vishwajeet Kumar, Yuncheng Hua, Ganesh Ramakrishnan, Guilin Qi, Lianli Gao, and Yuan-Fang Li. 2019.
\newblock Difficulty-controllable multi-hop question generation from knowledge graphs.
\newblock In \emph{The Semantic Web -- ISWC 2019}, pages 382--398, Cham. Springer International Publishing.

\bibitem[{Lewis et~al.(2020)Lewis, Liu, Goyal, Ghazvininejad, Mohamed, Levy, Stoyanov, and Zettlemoyer}]{bart}
Mike Lewis, Yinhan Liu, Naman Goyal, Marjan Ghazvininejad, Abdelrahman Mohamed, Omer Levy, Veselin Stoyanov, and Luke Zettlemoyer. 2020.
\newblock \href {https://doi.org/10.18653/v1/2020.acl-main.703} {{BART}: Denoising sequence-to-sequence pre-training for natural language generation, translation, and comprehension}.
\newblock In \emph{Proceedings of the 58th Annual Meeting of the Association for Computational Linguistics}, pages 7871--7880, Online. Association for Computational Linguistics.

\bibitem[{Li et~al.(2024)Li, Wang, Zhang, and Zhao}]{self-prompting}
Junlong Li, Jinyuan Wang, Zhuosheng Zhang, and Hai Zhao. 2024.
\newblock \href {https://doi.org/10.18653/v1/2024.naacl-long.17} {Self-prompting large language models for zero-shot open-domain {QA}}.
\newblock In \emph{Proceedings of the 2024 Conference of the North American Chapter of the Association for Computational Linguistics: Human Language Technologies (Volume 1: Long Papers)}, pages 296--310, Mexico City, Mexico. Association for Computational Linguistics.

\bibitem[{Li et~al.(2021)Li, Wang, and Yu}]{detecting-health-advice}
Yingya Li, Jun Wang, and Bei Yu. 2021.
\newblock \href {https://doi.org/10.18653/v1/2021.emnlp-main.486} {Detecting health advice in medical research literature}.
\newblock In \emph{Proceedings of the 2021 Conference on Empirical Methods in Natural Language Processing}, pages 6018--6029, Online and Punta Cana, Dominican Republic. Association for Computational Linguistics.

\bibitem[{Li et~al.(2023)Li, Zhu, Lu, and Yin}]{synthetic-roleplay}
Zhuoyan Li, Hangxiao Zhu, Zhuoran Lu, and Ming Yin. 2023.
\newblock \href {https://doi.org/10.18653/v1/2023.emnlp-main.647} {Synthetic data generation with large language models for text classification: Potential and limitations}.
\newblock In \emph{Proceedings of the 2023 Conference on Empirical Methods in Natural Language Processing}, pages 10443--10461, Singapore. Association for Computational Linguistics.

\bibitem[{Liu et~al.(2020)Liu, Chen, Liu, Bi, and Liu}]{ee-qa}
Jian Liu, Yubo Chen, Kang Liu, Wei Bi, and Xiaojiang Liu. 2020.
\newblock \href {https://doi.org/10.18653/v1/2020.emnlp-main.128} {Event extraction as machine reading comprehension}.
\newblock In \emph{Proceedings of the 2020 Conference on Empirical Methods in Natural Language Processing (EMNLP)}, pages 1641--1651, Online. Association for Computational Linguistics.

\bibitem[{Liu et~al.(2023)Liu, Wang, Yang, and Zha}]{financial-gpt}
Xiao-Yang Liu, Guoxuan Wang, Hongyang Yang, and Daochen Zha. 2023.
\newblock Data-centric fingpt: Democratizing internet-scale data for financial large language models.
\newblock \emph{NeurIPS Workshop on Instruction Tuning and Instruction Following}.

\bibitem[{Long et~al.(2024)Long, Wang, Xiao, Zhao, Ding, Chen, and Wang}]{synthetic-generation-survey}
Lin Long, Rui Wang, Ruixuan Xiao, Junbo Zhao, Xiao Ding, Gang Chen, and Haobo Wang. 2024.
\newblock \href {https://aclanthology.org/2024.findings-acl.658} {On {LLM}s-driven synthetic data generation, curation, and evaluation: A survey}.
\newblock In \emph{Findings of the Association for Computational Linguistics ACL 2024}, pages 11065--11082, Bangkok, Thailand and virtual meeting. Association for Computational Linguistics.

\bibitem[{MacQueen(1967)}]{kmeans}
J.~MacQueen. 1967.
\newblock \href {https://api.semanticscholar.org/CorpusID:6278891} {Some methods for classification and analysis of multivariate observations}.
\newblock In \emph{The Fifth Berkeley Symposium on Mathematical Statistics and Probability}.

\bibitem[{Markov et~al.(2023)Markov, Zhang, Agarwal, Eloundou~Nekoul, Lee, Adler, Jiang, and Weng}]{openai-guardrails}
Todor Markov, Chong Zhang, Sandhini Agarwal, Florentine Eloundou~Nekoul, Theodore Lee, Steven Adler, Angela Jiang, and Lilian Weng. 2023.
\newblock \href {https://doi.org/10.1609/aaai.v37i12.26752} {A holistic approach to undesired content detection in the real world}.
\newblock \emph{Proceedings of the AAAI Conference on Artificial Intelligence}, 37(12):15009--15018.

\bibitem[{Mazumder et~al.(2023)Mazumder, Banbury, Yao, Karla\v{s}, Gaviria~Rojas, Diamos, Diamos, He, Parrish, Kirk, Quaye, Rastogi, Kiela, Jurado, Kanter, Mosquera, Cukierski, Ciro, Aroyo, Acun, Chen, Raje, Bartolo, Eyuboglu, Ghorbani, Goodman, Howard, Inel, Kane, Kirkpatrick, Sculley, Kuo, Mueller, Thrush, Vanschoren, Warren, Williams, Yeung, Ardalani, Paritosh, Zhang, Zou, Wu, Coleman, Ng, Mattson, and Janapa~Reddi}]{data-perf}
Mark Mazumder, Colby Banbury, Xiaozhe Yao, Bojan Karla\v{s}, William Gaviria~Rojas, Sudnya Diamos, Greg Diamos, Lynn He, Alicia Parrish, Hannah~Rose Kirk, Jessica Quaye, Charvi Rastogi, Douwe Kiela, David Jurado, David Kanter, Rafael Mosquera, Will Cukierski, Juan Ciro, Lora Aroyo, and 28 others. 2023.
\newblock \href {https://proceedings.neurips.cc/paper_files/paper/2023/file/112db88215e25b3ae2750e9eefcded94-Paper-Datasets_and_Benchmarks.pdf} {Dataperf: Benchmarks for data-centric ai development}.
\newblock In \emph{Advances in Neural Information Processing Systems}, volume~36, pages 5320--5347. Curran Associates, Inc.

\bibitem[{Negi et~al.(2019)Negi, Daudert, and Buitelaar}]{semeval2019-task9}
Sapna Negi, Tobias Daudert, and Paul Buitelaar. 2019.
\newblock \href {https://doi.org/10.18653/v1/S19-2151} {{S}em{E}val-2019 task 9: Suggestion mining from online reviews and forums}.
\newblock In \emph{Proceedings of the 13th International Workshop on Semantic Evaluation}, pages 877--887, Minneapolis, Minnesota, USA. Association for Computational Linguistics.

\bibitem[{OpenAI et~al.(2024)OpenAI, Achiam, Adler, Agarwal, Ahmad, Akkaya, Aleman, Almeida, Altenschmidt, Altman, Anadkat, Avila, Babuschkin, Balaji, Balcom, Baltescu, Bao, Bavarian, Belgum, Bello, Berdine, Bernadett-Shapiro, Berner, Bogdonoff, Boiko, Boyd, Brakman, Brockman, Brooks, Brundage, Button, Cai, Campbell, Cann, Carey, Carlson, Carmichael, Chan, Chang, Chantzis, Chen, Chen, Chen, Chen, Chen, Chess, Cho, Chu, Chung, Cummings, Currier, Dai, Decareaux, Degry, Deutsch, Deville, Dhar, Dohan, Dowling, Dunning, Ecoffet, Eleti, Eloundou, Farhi, Fedus, Felix, Fishman, Forte, Fulford, Gao, Georges, Gibson, Goel, Gogineni, Goh, Gontijo-Lopes, Gordon, Grafstein, Gray, Greene, Gross, Gu, Guo, Hallacy, Han, Harris, He, Heaton, Heidecke, Hesse, Hickey, Hickey, Hoeschele, Houghton, Hsu, Hu, Hu, Huizinga, Jain, Jain, Jang, Jiang, Jiang, Jin, Jin, Jomoto, Jonn, Jun, Kaftan, Łukasz Kaiser, Kamali, Kanitscheider, Keskar, Khan, Kilpatrick, Kim, Kim, Kim, Kirchner, Kiros, Knight, Kokotajlo, Łukasz Kondraciuk,
  Kondrich, Konstantinidis, Kosic, Krueger, Kuo, Lampe, Lan, Lee, Leike, Leung, Levy, Li, Lim, Lin, Lin, Litwin, Lopez, Lowe, Lue, Makanju, Malfacini, Manning, Markov, Markovski, Martin, Mayer, Mayne, McGrew, McKinney, McLeavey, McMillan, McNeil, Medina, Mehta, Menick, Metz, Mishchenko, Mishkin, Monaco, Morikawa, Mossing, Mu, Murati, Murk, Mély, Nair, Nakano, Nayak, Neelakantan, Ngo, Noh, Ouyang, O'Keefe, Pachocki, Paino, Palermo, Pantuliano, Parascandolo, Parish, Parparita, Passos, Pavlov, Peng, Perelman, de~Avila Belbute~Peres, Petrov, de~Oliveira~Pinto, Michael, Pokorny, Pokrass, Pong, Powell, Power, Power, Proehl, Puri, Radford, Rae, Ramesh, Raymond, Real, Rimbach, Ross, Rotsted, Roussez, Ryder, Saltarelli, Sanders, Santurkar, Sastry, Schmidt, Schnurr, Schulman, Selsam, Sheppard, Sherbakov, Shieh, Shoker, Shyam, Sidor, Sigler, Simens, Sitkin, Slama, Sohl, Sokolowsky, Song, Staudacher, Such, Summers, Sutskever, Tang, Tezak, Thompson, Tillet, Tootoonchian, Tseng, Tuggle, Turley, Tworek, Uribe, Vallone,
  Vijayvergiya, Voss, Wainwright, Wang, Wang, Wang, Ward, Wei, Weinmann, Welihinda, Welinder, Weng, Weng, Wiethoff, Willner, Winter, Wolrich, Wong, Workman, Wu, Wu, Wu, Xiao, Xu, Yoo, Yu, Yuan, Zaremba, Zellers, Zhang, Zhang, Zhao, Zheng, Zhuang, Zhuk, and Zoph}]{GPT4.all}
OpenAI, Josh Achiam, Steven Adler, Sandhini Agarwal, Lama Ahmad, Ilge Akkaya, Florencia~Leoni Aleman, Diogo Almeida, Janko Altenschmidt, Sam Altman, Shyamal Anadkat, Red Avila, Igor Babuschkin, Suchir Balaji, Valerie Balcom, Paul Baltescu, Haiming Bao, Mohammad Bavarian, Jeff Belgum, and 262 others. 2024.
\newblock \href {https://arxiv.org/abs/2303.08774} {Gpt-4 technical report}.
\newblock \emph{Preprint}, arXiv:2303.08774.

\bibitem[{Park et~al.(2021)Park, Awadalla, Kohno, and Patel}]{distribution-shift-health}
Chunjong Park, Anas Awadalla, Tadayoshi Kohno, and Shwetak Patel. 2021.
\newblock \href {https://proceedings.neurips.cc/paper_files/paper/2021/file/17e23e50bedc63b4095e3d8204ce063b-Paper.pdf} {Reliable and trustworthy machine learning for health using dataset shift detection}.
\newblock In \emph{Advances in Neural Information Processing Systems}, volume~34, pages 3043--3056. Curran Associates, Inc.

\bibitem[{Rebedea et~al.(2023)Rebedea, Dinu, Sreedhar, Parisien, and Cohen}]{nemo}
Traian Rebedea, Razvan Dinu, Makesh~Narsimhan Sreedhar, Christopher Parisien, and Jonathan Cohen. 2023.
\newblock \href {https://doi.org/10.18653/v1/2023.emnlp-demo.40} {{N}e{M}o guardrails: A toolkit for controllable and safe {LLM} applications with programmable rails}.
\newblock In \emph{Proceedings of the 2023 Conference on Empirical Methods in Natural Language Processing: System Demonstrations}, pages 431--445, Singapore. Association for Computational Linguistics.

\bibitem[{Reddy et~al.(2017)Reddy, Raghu, Khapra, and Joshi}]{rnn-kg}
Sathish Reddy, Dinesh Raghu, Mitesh~M. Khapra, and Sachindra Joshi. 2017.
\newblock \href {https://aclanthology.org/E17-1036} {Generating natural language question-answer pairs from a knowledge graph using a {RNN} based question generation model}.
\newblock In \emph{Proceedings of the 15th Conference of the {E}uropean Chapter of the Association for Computational Linguistics: Volume 1, Long Papers}, pages 376--385, Valencia, Spain. Association for Computational Linguistics.

\bibitem[{Salem et~al.(2023)Salem, Paverd, and Köpf}]{salem2023maatphorautomatedvariantanalysis}
Ahmed Salem, Andrew Paverd, and Boris Köpf. 2023.
\newblock \href {https://arxiv.org/abs/2312.11513} {Maatphor: Automated variant analysis for prompt injection attacks}.
\newblock \emph{Preprint}, arXiv:2312.11513.

\bibitem[{Sennrich et~al.(2016)Sennrich, Haddow, and Birch}]{backtranslation}
Rico Sennrich, Barry Haddow, and Alexandra Birch. 2016.
\newblock \href {https://doi.org/10.18653/v1/P16-1009} {Improving neural machine translation models with monolingual data}.
\newblock In \emph{Proceedings of the 54th Annual Meeting of the Association for Computational Linguistics (Volume 1: Long Papers)}, pages 86--96, Berlin, Germany. Association for Computational Linguistics.

\bibitem[{Sudalairaj et~al.(2024)Sudalairaj, Bhandwaldar, Pareja, Xu, Cox, and Srivastava}]{instructlab}
Shivchander Sudalairaj, Abhishek Bhandwaldar, Aldo Pareja, Kai Xu, David~D. Cox, and Akash Srivastava. 2024.
\newblock \href {https://arxiv.org/abs/2403.01081} {Lab: Large-scale alignment for chatbots}.
\newblock \emph{Preprint}, arXiv:2403.01081.

\bibitem[{Taori et~al.(2020)Taori, Dave, Shankar, Carlini, Recht, and Schmidt}]{distribution-shift-images}
Rohan Taori, Achal Dave, Vaishaal Shankar, Nicholas Carlini, Benjamin Recht, and Ludwig Schmidt. 2020.
\newblock \href {https://proceedings.neurips.cc/paper_files/paper/2020/file/d8330f857a17c53d217014ee776bfd50-Paper.pdf} {Measuring robustness to natural distribution shifts in image classification}.
\newblock In \emph{Advances in Neural Information Processing Systems}, volume~33, pages 18583--18599. Curran Associates, Inc.

\bibitem[{Wang et~al.(2020)Wang, Cho, and Lewis}]{qa-summary}
Alex Wang, Kyunghyun Cho, and Mike Lewis. 2020.
\newblock \href {https://doi.org/10.18653/v1/2020.acl-main.450} {Asking and answering questions to evaluate the factual consistency of summaries}.
\newblock In \emph{Proceedings of the 58th Annual Meeting of the Association for Computational Linguistics}, pages 5008--5020, Online. Association for Computational Linguistics.

\bibitem[{Wang et~al.(2023)Wang, Kordi, Mishra, Liu, Smith, Khashabi, and Hajishirzi}]{self-instruct}
Yizhong Wang, Yeganeh Kordi, Swaroop Mishra, Alisa Liu, Noah~A. Smith, Daniel Khashabi, and Hannaneh Hajishirzi. 2023.
\newblock \href {https://doi.org/10.18653/v1/2023.acl-long.754} {Self-instruct: Aligning language models with self-generated instructions}.
\newblock In \emph{Proceedings of the 61st Annual Meeting of the Association for Computational Linguistics (Volume 1: Long Papers)}, pages 13484--13508, Toronto, Canada. Association for Computational Linguistics.

\bibitem[{Wang et~al.(2024)Wang, Li, Han, Nakov, and Baldwin}]{do-not-answer}
Yuxia Wang, Haonan Li, Xudong Han, Preslav Nakov, and Timothy Baldwin. 2024.
\newblock \href {https://aclanthology.org/2024.findings-eacl.61} {Do-not-answer: Evaluating safeguards in {LLM}s}.
\newblock In \emph{Findings of the Association for Computational Linguistics: EACL 2024}, pages 896--911, St. Julian{'}s, Malta. Association for Computational Linguistics.

\bibitem[{Xiang et~al.(2024)Xiang, Zheng, Li, Hong, Li, Xie, Zhang, Xiong, Xie, Yang, Song, and Li}]{guardagent}
Zhen Xiang, Linzhi Zheng, Yanjie Li, Junyuan Hong, Qinbin Li, Han Xie, Jiawei Zhang, Zidi Xiong, Chulin Xie, Carl Yang, Dawn Song, and Bo~Li. 2024.
\newblock \href {https://arxiv.org/abs/2406.09187} {Guardagent: Safeguard llm agents by a guard agent via knowledge-enabled reasoning}.
\newblock \emph{Preprint}, arXiv:2406.09187.

\bibitem[{Xu et~al.(2024)Xu, Cui, Yu, Kan, Shi, Zhuang, Wang, Jin, Ho, and Yang}]{synthetic-kg}
Ran Xu, Hejie Cui, Yue Yu, Xuan Kan, Wenqi Shi, Yuchen Zhuang, May~Dongmei Wang, Wei Jin, Joyce Ho, and Carl Yang. 2024.
\newblock \href {https://aclanthology.org/2024.findings-acl.916} {Knowledge-infused prompting: Assessing and advancing clinical text data generation with large language models}.
\newblock In \emph{Findings of the Association for Computational Linguistics ACL 2024}, pages 15496--15523, Bangkok, Thailand and virtual meeting. Association for Computational Linguistics.

\bibitem[{Ye et~al.(2022)Ye, Gao, Wu, Feng, Yu, and Kong}]{progen}
Jiacheng Ye, Jiahui Gao, Zhiyong Wu, Jiangtao Feng, Tao Yu, and Lingpeng Kong. 2022.
\newblock \href {https://doi.org/10.18653/v1/2022.findings-emnlp.269} {{P}ro{G}en: Progressive zero-shot dataset generation via in-context feedback}.
\newblock In \emph{Findings of the Association for Computational Linguistics: EMNLP 2022}, pages 3671--3683, Abu Dhabi, United Arab Emirates. Association for Computational Linguistics.

\bibitem[{Yoo et~al.(2021)Yoo, Park, Kang, Lee, and Park}]{synthetic-mixup}
Kang~Min Yoo, Dongju Park, Jaewook Kang, Sang-Woo Lee, and Woomyoung Park. 2021.
\newblock \href {https://doi.org/10.18653/v1/2021.findings-emnlp.192} {{GPT}3{M}ix: Leveraging large-scale language models for text augmentation}.
\newblock In \emph{Findings of the Association for Computational Linguistics: EMNLP 2021}, pages 2225--2239, Punta Cana, Dominican Republic. Association for Computational Linguistics.

\bibitem[{Yuan et~al.(2024)Yuan, Xiong, Zeng, Yu, Jia, Song, and Li}]{rigorllm}
Zhuowen Yuan, Zidi Xiong, Yi~Zeng, Ning Yu, Ruoxi Jia, Dawn Song, and Bo~Li. 2024.
\newblock \href {https://openreview.net/forum?id=QAGRPiC3FS} {Rigor{LLM}: Resilient guardrails for large language models against undesired content}.
\newblock In \emph{Forty-first International Conference on Machine Learning}.

\bibitem[{Zhang* et~al.(2020)Zhang*, Kishore*, Wu*, Weinberger, and Artzi}]{bertscore}
Tianyi Zhang*, Varsha Kishore*, Felix Wu*, Kilian~Q. Weinberger, and Yoav Artzi. 2020.
\newblock \href {https://openreview.net/forum?id=SkeHuCVFDr} {Bertscore: Evaluating text generation with bert}.
\newblock In \emph{International Conference on Learning Representations}.

\end{thebibliography}

\clearpage\appendix

\section{Performance on Real Data}\label{sec:appendix_bam_evaluation}
We recognize that given the significant distribution shift between fine-tuning data and real-world production data, performance on task evaluation benchmarks may not necessarily transfer over in practice. Thus, we validate the performance of our model on internal real-world production data, which consists of 5k samples. Of these 5k samples, 4642 are general content, 350 are health related (but not advice), and only 8 are health advice. Note that this vastly skewed label distribution is actually normal, since health advice composes a very minute portion of all the real-world chatlogs. 

\begin{table}[htbp]
\begin{center}
\small
\begin{tabular}{|c|c|c|}
\hline
FT Strategy & Accuracy $\uparrow$ & FPR $\downarrow$ \\
\hline\hline 
Detector 2-Stage & 97.14\% & 2.86\% \\
Detector 1-Stage & \textbf{99.24}\% & \textbf{0.70}\% \\
Vanilla Academic & 96.10\% & 3.88\% \\
Vanilla Synthetic & 97.54\% & 2.46\% \\
\hline
\end{tabular}
\end{center}
\caption{A comparison of performance on 5k samples of real production data. Note that academic data refers to the training dataset where we use the original seed datasets instead of the synthetic data. Synthetic data refers to using only the synthetically generated data. Note that FPR stands for false positive rate. The best performing results are in \textbf{bold}.}
\label{tab:bam_eval_results}
\end{table}

From our results in Table~\ref{tab:bam_eval_results}, we see that our best performing setup is just after one stage of FT with a mixture of synthetic and academic data, achieving by far the best false positive rate of just 0.70\% (statistically significant to 99\% confidence compared to the next lowest rate at 2.46\%), and an accuracy of 99.24\%. We posit that this is due to the model seeing the widest range of samples, and thus is more robust when encountering these samples out in the real world. Interestingly, two stage FT performs comparably with vanilla FT using only synthetic data (with the difference in results not statistically significant), suggesting that in the real-world, it may be more beneficial to go directly towards the LLM output distribution by directly FT on purely synthetic data. However, we note that this subset of real-world data was taken from a contiguous time period, and thus may not be fully representative of the real-world. Nevertheless, the benefits of training on synthetic data cannot be understated, as FT on purely academic data results in a notable degradation in real-world performance, with its error rate of 3.88\% being statistically significant (99\% confidence) compared to the next highest rate at 2.86\%. 

\section{Dataset Statistics}\label{sec:appendix_datasets}
The HeAL evaluation benchmark contains a total of 402 samples, of which 241 constitute health advice and 161 are negative samples. Detecting-Health-Advice contains 10848 samples, of which 2747 constitute health advice. HealthE contains 5656 samples, of which 3400 constitute health advice. SemEval2019-Task9 contains 9925 samples, all of which are labeled general content (i.e. negative samples). 

The fine-tuning dataset for the first stage contains 26429 total samples, of which 10356 constitute health content (negative samples), 6148 constitute health advice, and 9925 constitute general content (also negative samples). The fine-tuning dataset for the second stage contains 6147 total samples, of which 2916 constitute health content (negative samples), 2343 constitute health advice, and 888 constitute general content.

\section{Hyperparameters}\label{sec:appendix_hyperparameters}
In the backprompting framework, we used the same hyperparameters for both query generation and output generation. We set the minimum number of new tokens to be 5, and a maximum amount of new tokens to be 250. We sample with a temperature of 0.6, renormalize logits, and furthermore restrict a no repeat n-gram size of 5. 
All FT stages utilize the same hyperparameters, which are relatively standard. We use a learning rate of 2e-5, a batch size of 16, FT for 5 epochs, and use a weight decay regularization parameter of 0.01. 

\section{Software \& Model Implementation}\label{sec:appendix_software_model_compute}
Our implementation is written in Python, using PyTorch and Huggingface's Transformers library. Our framework is readily implementable on as little as 1 V100 GPU with 32 GB of GPU memory. For the larger LLMs, we load them for generation using 4-bit quantization. FT experiments and simulations can be executed in just a few hours, and typically less than half a day. As a result, we expect our environmental and carbon emissions impact to be relatively low-cost. 

\section{Prompt \& Output Examples}\label{sec:appendix_examples}
The prompt template for the backprompting framework is shown in Table~\ref{tab:star_prompt_generation}. We also show three examples of the synthetic data, paired with the queries generated via backprompting, as well as the original seed text. These examples are displayed in Table~\ref{tab:qualitative_analysis}.

\begin{table}[t]
    \centering
    \begin{tabular}{|p{0.95\linewidth}|}
      \hline \textbf{Prompt} \\ \hline
      ``What question did the user ask to generate the following text:\newline\newline\{$x_i$\}\newline\newline The user prompt is:"
      \\ \hline
    \end{tabular}
    \caption{The prompt that we used to generate queries in our framework. Note that $x_i \in \mathcal{X}$ represents one sample from our seed dataset.}
    \label{tab:star_prompt_generation}
\end{table}

\begin{table*}[htbp]
    \centering
    \begin{tabular}{|p{57mm}|p{27mm}|p{57mm}|}
      \hline \textbf{Original} & \textbf{Prompt} & \textbf{Synthetic Output} \\ \hline
      Georgian public health specialists working in the HIV field should prioritize implementation of such interventions among HIV patients. & What should Georgian public health specialists working in the HIV field prioritize? & The Georgian Public Health Association (GPHA) is organizing a conference, “HIV in Georgia: From Research to Practice,” to address the challenges of HIV in the country. Public health specialists working in the HIV field should prioritize the following: (1) Strengthening the healthcare system: The conference should focus on the capacity of the healthcare system to provide comprehensive HIV services, including testing, treatment, and care. This includes ensuring that healthcare providers have the necessary training, equipment, and resources to deliver high-quality services... \\ \hline
      Continued surveillance is needed to monitor the changing epidemiology of CA-MRSA and to inform clinical decisions. & What are the benefits of supporting patients with oral chemotherapy programs? & Oral chemotherapy programs have been shown to improve patient outcomes by increasing adherence to treatment, reducing hospitalization rates, and enhancing quality of life. Here are some of the benefits of supporting patients with oral chemotherapy programs: (1) Improved Adherence to Treatment: Oral chemotherapy programs can improve patient adherence to treatment by providing a convenient and flexible option for medication administration. This can lead to better disease outcomes and improved survival rates... \\ \hline
      Findings provide preliminary support that rosiglitazone may offer a novel strategy for the treatment of cognitive decline associated with AD. & What are the advantages of expressing plasma FA as absolute concentrations over relative units? & The advantages of expressing plasma FA as absolute concentrations over relative units include: (1) Improved accuracy: Absolute concentrations provide a more accurate representation of the amount of fatty acids in the plasma, as they are not affected by variations in total cholesterol levels. (2) Enhanced sensitivity: Absolute concentrations can detect smaller changes in fatty acid levels, making it easier to identify subtle changes in the plasma lipid profile...\\ \hline
    \end{tabular}
    \caption{We compare the original seed examples with their generated prompts and corresponding synthetic outputs. For brevity, the synthetic outputs shown for these examples have been truncated to fit on this page.}
    \label{tab:qualitative_analysis}
\end{table*}

\end{document}